\documentclass[runningheads]{llncs}
\usepackage[T1]{fontenc}
\usepackage{graphicx}
\usepackage{booktabs}						
\usepackage{multirow}						
\usepackage{amsfonts}						
\usepackage{graphicx}						
\usepackage{duckuments}						
\usepackage{graphicx}
\usepackage{colortbl,xcolor}
\usepackage{tikz}
\usepackage{amsmath}
\usepackage{hyperref}
\usetikzlibrary{arrows.meta, positioning, fit}
\usetikzlibrary{shapes.geometric, arrows.meta, positioning}
\usepackage{array} 
 \usepackage{amssymb}
\usepackage{booktabs}
\usepackage{graphicx}
\usepackage{tikz}
\usetikzlibrary{arrows.meta, positioning, fit}
\usetikzlibrary{shapes.geometric, arrows.meta, positioning}
\usepackage{array} 
 \usepackage{amssymb}
 \usepackage{tocloft} 

\usepackage{booktabs}
\usepackage{listings}
\usepackage{xcolor}
\usepackage{array} 

\usepackage{booktabs}
\usepackage{multirow}
\def\eg{\emph{e.g}}

\usepackage{amssymb} 
\newcommand{\cmark}{\checkmark} 
\newcommand{\xmark}{\text{\sffamily X}} 

\usepackage{adjustbox} 
\usepackage{adjustbox} 
\lstdefinestyle{modernstyle}{
    backgroundcolor=\color{white},      
    commentstyle=\color{gray!50!black}\itshape,  
    keywordstyle=\color{blue!80!black}\bfseries, 
    numberstyle=\tiny\color{gray},      
    numbers=left,                       
    stepnumber=1,                       
    numbersep=10pt,                     
    stringstyle=\color{purple!70!black}, 
    basicstyle=\ttfamily\small,         
    breaklines=true,                    
    captionpos=b,                       
    showspaces=false,                   
    showstringspaces=false,             
    showtabs=false,                     
    frame=single,                       
    rulecolor=\color{gray!40!black},    
    tabsize=2,                          
    linewidth=0.9\textwidth,            
    xleftmargin=10pt,                   
    xrightmargin=10pt,                  
    aboveskip=1em,                      
    belowskip=1em,                      
    escapeinside={\%*}{*},              
    caption={},                         
}
\newcolumntype{C}[1]{>{\centering\arraybackslash}m{#1}}

\usepackage{biblatex}
\begin{document}
\title{Visual Model Checking:\\Graph-Based Inference of Visual Routines for Image Retrieval}
\titlerunning{Visual Model Checking}
%
\author{Adrià Molina \and Oriol Ramos Terrades \and Josep Lladós}
\authorrunning{A. Molina et al.}
%
\institute{Centre de Visió per Computador \and Universitat Autònoma de Barcelona \\   \email{\{amolina, oriolrt, josep\}@cvc.uab.cat}}
\maketitle              
\begin{abstract}
Information retrieval lies at the foundation of the modern digital industry. While natural language search has seen dramatic progress in recent years largely driven by embedding-based models and large-scale pretraining, the field still faces significant challenges. Specifically, queries that involve complex relationships, object compositions, or precise constraints such as identities, counts and proportions often remain unresolved or unreliable within current frameworks. In this paper, we propose a novel framework that integrates formal verification into deep learning–based image retrieval through a synergistic combination of graph-based verification methods and neural code generation. Our approach aims to support open-vocabulary natural language queries while producing results that are both trustworthy and verifiable. By grounding retrieval results in a system of formal reasoning, we move beyond the ambiguity and approximation that often characterize vector representations. Instead of accepting uncertainty as a given, our framework explicitly verifies each atomic truth in the user query against the retrieved content. This allows us to not only return matching results, but also to identify and mark which specific constraints are satisfied and which remain unmet, thereby offering a more transparent and accountable retrieval process while boosting the results of the most popular embedding-based appoaches.

\end{abstract}

\section{Introduction}
\label{sec:intro}



Image search plays a foundational role in modern digital ecosystems, serving as a critical interface between users and vast repositories of visual content. As embedding-based retrieval systems have grown more capable, they have also become more relied upon for interpreting complex scenes and satisfying open-ended queries. However, these systems typically prioritize approximation over precision, offering plausible results without guaranteeing that specific elements of a query are actually satisfied. This approximation, while efficient for routine tasks, becomes problematic when combined with inadvertent misinformation \cite{epstein2024can} or deliberate manipulation of content~\cite{ludolph2016manipulating}.

Our work is a response to this fundamental tension between flexibility and reliability in image retrieval. By integrating formal verification into the retrieval process, each component of a natural language query,represented as logical triplets,can be independently tested against candidate images. This enables the system to explicitly signal which constraints are satisfied, which are violated, and which remain indeterminate. In high-stakes scenarios where factual accuracy is crucial, such verifiability transforms visual search from an opaque process of approximation into a transparent mechanism for structured reasoning,anchoring results in logic rather than intuition or heuristic match.

\begin{figure}[t]
    \centering
\begin{tabular}{ccccc}
\multicolumn{2}{c}{Specification}  & \multicolumn{2}{c}{Description}      &                              \\
\toprule
Query-to-Graph & Graph-to-Code & Candidate Images & Execute and Prune & 
\multirow{3}{*}{\raisebox{-0.5\height}{\includegraphics[]{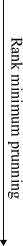}}} \\
\raisebox{-0.5\height}{\includegraphics[]{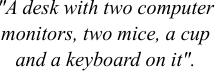}} &
\raisebox{-0.5\height}{\includegraphics[]{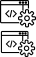}} &
\raisebox{-0.5\height}{\includegraphics[width=.17\textwidth]{images/visual_abstract/composicio/000000148620.jpg}} &
\raisebox{-0.5\height}{\includegraphics[]{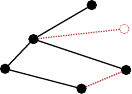}} &
\\
\raisebox{-0.5\height}{\includegraphics[width=0.25\textwidth]{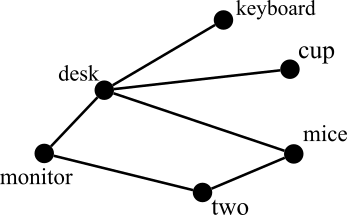}} &
\raisebox{-0.5\height}{\includegraphics[]{images/visual_abstract/composicio/path1.png}} &
\raisebox{-0.5\height}{\includegraphics[width=.17\textwidth]{images/visual_abstract/composicio/000000371677.jpg}} &
\raisebox{-0.5\height}{\includegraphics[]{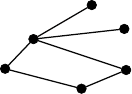}} &
\end{tabular}

\caption{Visual abstract of the proposed visual search framework. The system follows a model-checking approach: the textual query is first converted into a structured system specification (graph). For each local specification, a corresponding visual routine is generated (code). Candidate images are then ranked based on how well they satisfy these visual routines (prunning), enabling partial or complete verification of visual content in digital environments.}
    \label{fig:visual_abstract}
\end{figure}

In this manuscript, we introduce a methodology inspired by the model checking framework~\cite{clarke2018introduction}, which serves as a prominent example in formal verification. As depicted in \figureautorefname~\ref{fig:visual_abstract}, a natural language query is interpreted as a \textit{system specification}. Based on this interpretation, a world model is constructed and, then, compared against specific instances of that model. In our approach, the query defines the desired specifications for the world model, which can be instantiated through what we call a \textit{visual routine}—a visual grammar that abstracts structures from particular instantiations of how the world should look. In essence, this enables the retrieval of information while detecting partial truths, introducing a regime where doubt is incorporated through contradictions between the declared specifications and the visual routines observed.

Therefore, the contributions of this work are:
\begin{itemize}

\item A formal visual grammar framework is introduced that adapts model checking techniques to image retrieval tasks, offering additional verification guarantees.

\item We present a system that achieves competitive performance with state-of-the-art visual search methods, thereby validating the effectiveness of the proposed framework.

\item We provide an extensive qualitative evaluation that highlights key advantages and common pitfalls of our approach across diverse visual search scenarios.
\end{itemize}

The rest of the work is organized as follows: We introduce a brief literature review in Section~\ref{sec:sota}. Methodology is explained in Section~\ref{sec:method} and experimental setup in Section~\ref{sec:exp}. Results are presented in Section~\ref{sec:results} and conclusions in Section~\ref{sec:conc}. 

\section{Related work}
\label{sec:sota}
This paper lies at the intersection of image search and code-based methods. Accordingly, in this section, we review recent advancements and challenges in both areas. Our proposed method operates at the convergence of these domains, aiming to offer new perspectives on image search through a joint approach.

In the recent years image search has been dominated by embedding-based self-supervised multi-modal image encoders~\cite{radford2021learning,xu2023metaclip,wang2023image,jia2021scaling}. Although early stages of bridging visual descriptors and topic modeling can be dated back to 2017~\cite{gomez2017self}, the introduction of large-scale pretraining in CLIP~\cite{radford2021learning} took over the whole visual search landscape and multiple variants came through the following years~\cite{zhai2023sigmoidlosslanguageimage}.

As noted in ~\cite{wang2024balanced}, embedding-based methods often struggle with textual queries that involve multi-faceted or compositional descriptions. In this context, probabilistic approaches have shown promising improvements~\cite{chun2021probabilistic, chun2023improved}. However, these methods still face challenges when it comes to reasoning over structured information and handling complex operations: arithmetic, geometric localization and text recognition.

\begin{figure}[t]
    \centering
    \resizebox{\textwidth}{!}{%
    \begin{tikzpicture}[
        node distance=1.2cm and 1.8cm,
        every node/.style={align=center},
        box/.style={draw, rounded corners, text width=3.5cm, minimum height=1cm, text centered},
        arrow/.style={-{Stealth}, thick},
        ]

                \node[box] (spec) {system specification\\ (query $q$)\\};

        \node[box, below=of spec.center] (desc) {system description\\ (image $v$) \\};

        \node[box, right=of spec] (parse) {system parser\\ $P(q) = \varphi$};
        \node[box, below=of parse.center] (check) {visual verifier\\ checks if $\pi_i(v) = \text{True}$};

        \node[box, right=of parse] (model) {Visual routines $\Pi$\\ (world model)\\ $\pi_i = \text{M}(\varphi_i)$};

        \node[box, below=of check, text width=10cm] (output) {
            \begin{itemize}
                \item If all $\pi_i(v) = \text{True}$: \textit{``$v$ satisfies $\varphi$''}
                \item Else: \textit{``visual counterexample to $\varphi$ in $v$''}
            \end{itemize}
        };

        \draw[arrow] (spec) -- node[above] {P}  (parse);
        \draw[arrow] (desc) -- (check);
        \draw[arrow] (parse) -- node[above] {M} (model);
        \draw[arrow] (model.south) |- (check.east);
        \draw[arrow] (check) -- (output);

\node[draw, dashed, fit=(spec)(desc)(parse)(model)(check), inner sep=0.7cm, 
      label={[below]: \textbf{Model Checking with Visual Routines}}] (container) {};

    \end{tikzpicture}}
    \caption{Layout of our model-checking methodology with visual routines heavily inspired by \cite{clarke2018introduction}.}
    \label{fig:model_checking}
\end{figure}
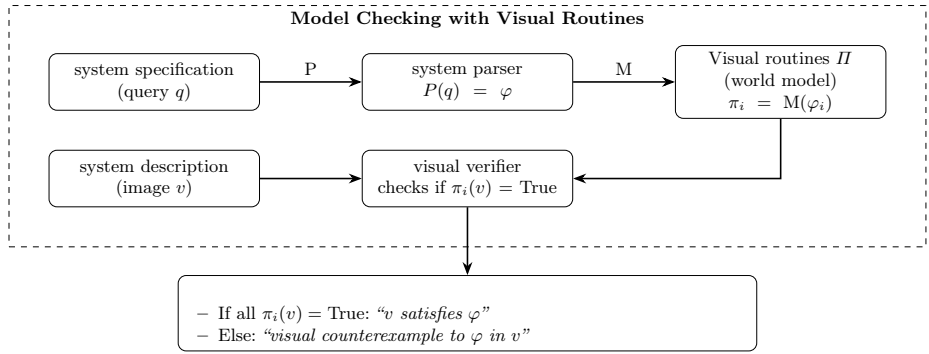

Recent work at the intersection of vision and symbolic reasoning has attempted to bridge high-level semantic understanding with visual perception through program synthesis. Two representative approaches are the neuro-symbolic framework Pix2Code and the code-based reasoning system ViperGPT.

Pix2Code~\cite{wust2024pix2code} introduces a neuro-symbolic framework that learns to compose visual concepts as executable programs. It combines neural representations of objects with symbolic reasoning, enabling generalizable and revisable concept learning. However, the system primarily functions as a one-way encoder: given an image, it generates a program that describes its contents. This design limits its applicability in retrieval tasks, as it is oriented toward query-by-example rather than query-by-specification. Furthermore, Pix2Code lacks scalability in open-vocabulary settings and, as noted by the authors, its capabilities are largely confined to simple visual properties such as object counts and spatial arrangements. More complex semantics,such as actions seen in Visual Genome~\cite{krishna2017visual} or nuanced inter-object relationships (\eg, ``person feeding a dog''),remain beyond its scope.


Both ViperGPT~\cite{suris2023vipergpt} and VisProg~\cite{gupta2023visual} introduces a zero-shot framework that answers visual queries by generating Python programs composed of API calls to pretrained perception models (\eg, object detectors, attribute classifiers). Rather than reasoning visually, it defines the logic of visual checks and delegates execution, enabling interpretability, compositionality, and strong generalization without further training. Building on this idea, we apply formal verification principles,commonly used in program analysis,to visual retrieval. Like ViperGPT, we use language models\footnote{In contrast to \cite{suris2023vipergpt} and \cite{gupta2023visual}, which rely on proprietary models, our work leverages open-source LLMs, thereby facilitating reproducibility and contributing to the community in an open and transparent manner.} to generate structured specifications and offload low-level tasks to specialized routines. However, inspired by work on interpretable world models~\cite{ellis2021dreamcoder,wust2024pix2code}, our goal is not to answer specific queries, but to enable retrieval by ranking images according to how well they satisfy a reusable set of logical triplets.

\section{Methodology}
\label{sec:method}

Our approach is inspired by model checking~\cite{clarke2018introduction}. As in that domain, we define the system components illustrated in Figure~\ref{fig:model_checking}. The following sections introduce the required notation and describe each element,from interpreting the user query as a specification to generating the final retrieval output.

\subsection{Notation}
Our proposed framework is constructed under the following elements:

\begin{enumerate}
    \item A \textit{system specification} expressed as a natural language query $q$, which can be formulated in an open-vocabulary manner.

    \item A \textit{system description} represented by a visual element $v$ (an image) that serves as a potential instantiation of the \textit{system specification}.
    
\item A \textit{system parser} function $P(q) = \varphi$ which transforms user queries into a graph structure: $\{(s_1, p_1, o_1), (s_2, p_2, o_2), \ldots, (s_n, p_n, o_n)\} $.

This graph encodes the specification as a set of subject-predicate-object triples $\varphi_i$, parsed independently of any image, as $\varphi$ defines the target conditions for verification.

    \item A \textit{function synthesizer} model, that maps $\varphi_i$ triples to its associated routine $\pi_i$:
    \begin{equation*}
            \text{M}: \varphi \rightarrow \Pi
    \end{equation*}
    \item A set of \textit{visual routines} $\Pi$, consisting of specialized program $\pi_i$ that satisfy the triplets of $\varphi$. Formally:
    \begin{equation*}
            \pi_i: V \rightarrow \{\text{True, False}\}
    \end{equation*}
    \begin{equation*}
        \Pi = \{\pi_i \mid \pi_i \text{ is a visual routine corresponding to } (s_i, p_i, o_i) \in \varphi\}
    \end{equation*}
    where each $\pi_i = \text{M}(\varphi_i)$ is a program that captures the triple $(s_i, p_i, o_i)$ into executable visual grammars and determines its validity in the visual domain.

\end{enumerate}
Therefore, under this notation, a specification $\varphi$ is first defined. This is ultimately mapped to $\Pi$, where each element is a verification function. An image $v$ satisfies $\varphi$ (and is therefore verified by the world model $\Pi$) if and only if every element $\pi_i \in \Pi$ evaluates to True when applied to $v$.

\subsection{Retrieval method}
Since $\Pi$ maps images to a set of boolean outputs, this framework provides a binary verification mechanism: whether an image satisfies a given specification. However, in the context of image retrieval, this raises a challenge,how can such a model be used to rank images, rather than merely verifying them?

\begin{figure}[t]
    \includegraphics[width=\textwidth]{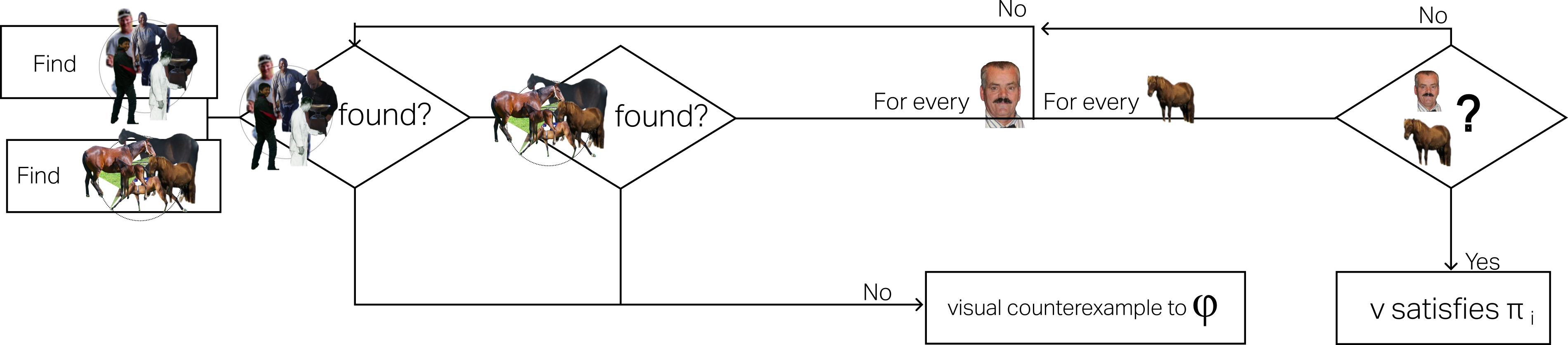}
    \caption{Example of an actual visual routine for $(man, riding, horse)$, on which a program is synthesized to recognize an abstract statement with agnosticity of any specific image.}
    \label{fig:visual_routine}
\end{figure}

\paragraph{Ranking score:} We introduce a ranking mechanism for image retrieval by extending our Boolean verification framework through partial validation. Since each visual routine~$\pi_i$ evaluates a specific aspect of a specification, we can rank images by calculating the proportion of routines they satisfy. This generates a truth score that quantitatively reflects an image's compliance with the original query, enabling effective ranking rather than mere binary verification.


\paragraph{Visual Routines as Re-Rankers}{In the previous steps, our method would output a score between 0 and 1 proportional to the fraction of verified routines from a query. With this, one can weight the original retrieval score to perform a re-ranking. In our case; we select the top-k ranked images to obtain a re-rank from $K$; and create a score $K-i$ for each image retrieved as $i$-th element of the ranking (hence, obtaining a retrieval score). We weight this score by the previously obtained fraction:}
\begin{equation}
    \text{ReRankScore}_i = (K - i) \times \frac{\text{\# Verified Triplets}}{\text{\# Total of Triplets}}
\end{equation}

This hybrid scheme between symbolic and embedding-based approaches is also evaluated and proved worthy of boosting results in some embedding-based models.


\paragraph{Granularity level:} Note that one can sample many subgraphs from a set of constraints, up to a maximum determined by the cartesian product of $\varphi$. This combinatorial growth is known as state-explosion. Our proposed approach constructs global conclusions by aggregating local truths from individual validations: An statement is true, if it is composed by true statements~\cite{giannakopoulou2018compositional}. While global reasoning cannot be completely validated through triplets alone in all cases, this assumption effectively mitigates the combinatorial growth of possible states,a phenomenon directly analogous to the state explosion problem in classical model checking.

\paragraph{Pipeline:} Our workflow, following \figureautorefname~\ref{fig:model_checking}, begins with parsing a user query $q$ into a logical specification $\varphi$, consisting on subject-predicate-object triplets. In order to avoid state-explosion, each triplet forms the basis for a visual routine~$\pi$, with our secondary model, $\text{M}(\varphi)$, transforming these logical triplets into executable programs. As illustrated in Figure~\ref{fig:visual_routine}, this process synthesizes routines capable of detecting specific visual relationships (\eg, ``men riding horses'') within images.

\paragraph{Implementation of the pipeline:} Following prior literature~\cite{suris2023vipergpt}, these visual routines are synthesized through a large language model with access to a predefined modular API. For each triplet in the logical specification $\varphi$, the LLM generates a corresponding visual routine defining necessary conditions for satisfaction. These routines are then executed using open-vocabulary visual detection models, enabling flexible verification across diverse concepts. This approach translates symbolic specifications into executable visual checks that can be efficiently evaluated across candidate images. The images are then ranked based on the number of satisfied routines,effectively producing an ordering derived from the extent to which each image fulfills the specification.

In the following sections, readers can access to the implementation details of the used models to perform the task, and how this approach is validated.

\section{Experimental Set-Up}
\label{sec:exp}

This manuscript presents a central claim that has yet to be fully validated. Namely, that the proposed model-checking approach holds promising potential for improving visual search algorithms. In this section, we describe the evaluation methodology and implementation setup used to validate this claim.

\subsection{Data and Evaluation}
\begin{figure}[t]
    \centering
    \renewcommand{\arraystretch}{1.2} 
    \begin{tabular}{p{0.22\textwidth} p{0.22\textwidth}|p{0.22\textwidth} p{0.22\textwidth}}
    \multicolumn{2}{c}{CoCo-Easy} & \multicolumn{2}{c}{CoCo-Hard} \\
    \includegraphics[width=\linewidth]{images/coco_examples/000000327780.jpg} & 
    \includegraphics[width=\linewidth]{images/coco_examples/000000122672.jpg} & 
    \includegraphics[width=\linewidth]{images/coco_examples/000000157098.jpg}&
    \includegraphics[width=\linewidth]{images/coco_examples/000000251119.jpg} \\
    \textit{A fork cutting into a piece of cake.} &
    \textit{A guy on a surf board in the water.} &
    \textit{Two giraffes make contact while a third eats from a tree.} &
    \textit{ Boxed meal of sandwich roll, orange juice and strawberry yogurt} \\
    \end{tabular}
    
\caption{Examples of easy and hard instances from the MS-COCO dataset~\cite{lin2014microsoft}. Qualitatively, harder examples tend to involve more intricate descriptions, including object counting, text recognition, and reasoning about relationships between entities.}
    \label{fig:caption_splits}
\end{figure}

We aim to demonstrate the potential of our framework in a controlled text-to-image retrieval setting using the MS-COCO Captions dataset~\cite{lin2014microsoft}, specifically its 2017 validation split.

Note that not all MS-COCO captions involve the type of complex reasoning required to test our research hypothesis~\cite{stefanini2022show,sarto2025image}. Therefore, it is necessary to examine a subset of examples that both support this validation and ensure a fair evaluation across a diverse range of scenarios.

To address this evaluation bias, we partition the validation set into COCO-Easy and COCO-Hard, based on CLIP’s~\cite{radford2021learning} retrieval performance (see  \figureautorefname~\ref{fig:caption_splits}). The Easy split comprises the top 25\% of pairs where CLIP excels, while the Hard split includes the bottom 25\%, where it struggles,typically in cases involving compositionality, text recognition, or counting. The whole COCO dataset is also evaluated and noted as ``COCO-All'' for clarity.

We evaluate the accuracy of our method on each split individually and jointly (noted as combined), comparing against other SOTA embedding-based models. Results show consistent trends across models, offering a fairer performance assessment.
\subsection{Implementation Details}

Our system is deployed in a distributed multi-node environment with heterogeneous GPU resources, including nodes with 4$\times$A40, 2$\times$A40, and 4$\times$RTX 3090 GPUs. It follows a producer-consumer architecture, where a central producer manages task scheduling, and consumer processes execute different pipeline stages based on available hardware. The pipeline consists of three stages: parsing natural language queries into subject-predicate-object triplets using Microsoft Phi-4~\cite{abdin2024phi}; synthesizing Python-based visual routines from logical graphs via the same model; and executing these routines along OWL-V2~\cite{minderer2023scaling} to perform symbolic verification on candidate images.

Visual inference is assigned to high-performance nodes, while lightweight tasks are broadly distributed to maximize throughput. All experiments use fixed seeds for reproducibility, and we will release the full codebase, and evaluation tools for the community.
 
\section{Results and Discussion}
\label{sec:results}



\paragraph{Quantitative Evaluation}{As shown in Table~\ref{tab:resultats_quantitatius}, our method (alone) achieves competitive performance in retrieval recall at ranks 1, 5, and 10, matching state-of-the-art methods at higher K cutoffs. This trend reverses in the COCO-Hard split, where our approach demonstrates clear advantages, attributable to the model ability of the approach to verify information involving textual content, spatial composition, and basic numerical reasoning. The combined results (Easy+Hard) highlight a key limitation of our approach: inaccuracies in code synthesis and failures in the visual detection API introduce noise that partially offsets the gains achieved on COCO-Hard.

When evaluating our method in conjunction to CLIP, BEIT and ALIGN we observe a very positive trend: While one might be tempted to infer that there is no winning solution, using both methods jointly proofs worthy in most of the situations, as the performance of embedding-based methods is improved in each and every evaluation scenario.

\begin{table}[t]
\caption{Recall for our approach and other embedding-based methods We observe competitive performance to every zero-shot method. Note that, as can be seen in the qualitative evaluation, most of COCO captions could be attributed to a wide set of images. In contrast to the rest of comparisons, including ours, BEIT-3 \cite{wang2023image} results are not zero-shot.}
\centering
     \centering
\resizebox{\textwidth}{!}{
\begin{tabular}{@{}lccccccccc c@{}}
\toprule
                           & \multicolumn{3}{c}{Easy} & \multicolumn{3}{c}{Combined} & \multicolumn{3}{c}{Hard} & Zero-Shot? \\ \midrule
                           \textbf{Recall:} & Rec@1 & Rec@5 & Rec@10 & Rec@1 & Rec@5 & Rec@10 & Rec@1 & Rec@5 & Rec@10 &  \\ \midrule
CLIP \tiny{(ViT-B)}~\cite{radford2021learning} & .486 & .850 & .892 & .395 & .710 & .884 & .152 & .430 & .565 & \cmark \\
CLIP \tiny{(ViT-H, QuickGELU)}~\cite{xu2023metaclip} & .\textbf{649} & .919& .973 & .605 & .899& .961 & .283 & .413& .674 & \cmark \\
SigLIP \tiny{(400M)}~\cite{zhai2023sigmoidlosslanguageimage} & .568 & .811 & .892 & .581 & .853 & .946 & .304 & .478& .696 & \cmark \\
BEIT-3 \tiny{(COCO-Finetuned)}~\cite{wang2023image} & .676 & \textbf{.959}& .973 & \textbf{.752} & \textbf{.944}& \textbf{.984} & .\textbf{435} & \textbf{.739}& .870 & \xmark \\
ALIGN \tiny{\cite{jia2021scalingvisualvisionlanguagerepresentation}} & .459 & .469& .838 & .481 & .827& .938 & .283 & .413& .565 & \cmark \\ \midrule
\multicolumn{1}{l}{(ours, base)} & .432 & .550 & .784 & .450 & .700 & .868 & .196 & .490 & .739 & \cmark \\ 
\multicolumn{1}{l}{(ours) + CLIP} & .592 & \textbf{.959} & \textbf{.980} & .429 & .638 & .679 & .043 & .261 & .565 & \cmark \\ 
\multicolumn{1}{l}{(ours) + BEIT} & .551 & .939 & \textbf{.980} & .689 & .913 & .969 & .391 & .717 & \textbf{.913} & \xmark \\ 
\multicolumn{1}{l}{(ours) + ALIGN} & .469 & .857 & .878 & .556 & .872 & .934 & .261 & .543 & .717 & \cmark \\ 
\bottomrule
\end{tabular}

}

\label{tab:resultats_quantitatius}
\end{table}
Additionally, our method on its own remains competitive overall and shows clear advantages in scenarios involving complex captions,particularly those requiring reasoning about relationships, composition, and numerical attributes,while also offering improved verification and reliability compared to baseline methods.}

\begin{figure}
    \centering
    \includegraphics[width=1\linewidth]{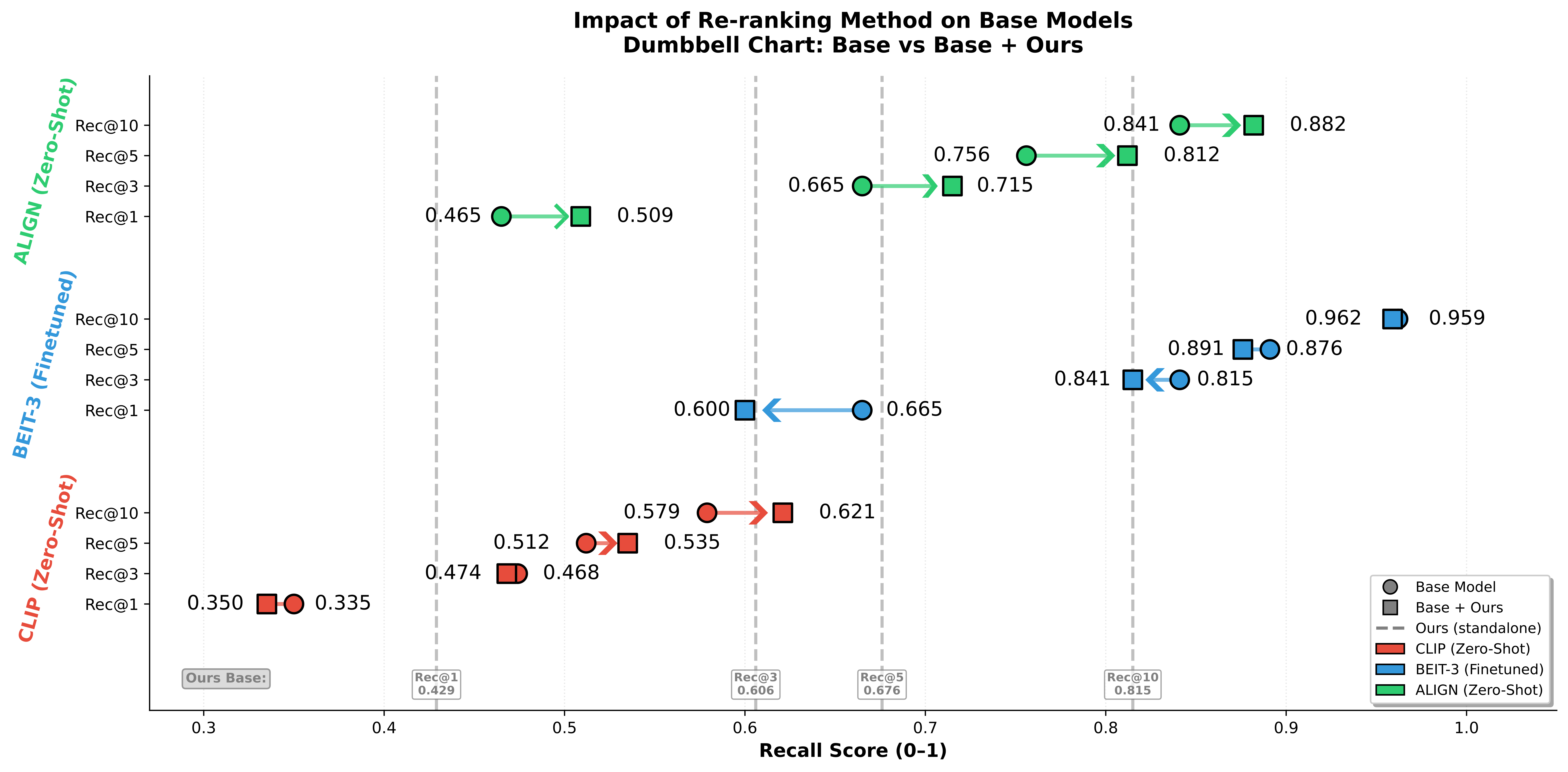}
    \caption{Re-ranking embedding-based content (circles) with our method (squares).}
    \label{fig:reranking_effect}
\end{figure}

Additional results computed in COCO-All are presented in Table~\ref{tab:coco_all_results}, which provide comparison of retrieval performance between our method and state-of-the-art embedding-based approaches. In Figure~\ref{fig:reranking_effect} we observe how the usage of visual routines in re-ranking embedding-based queries improves the performance in many zero-shot scenarios. The drop in performance in fine-tuned instances (BEIT-3, blue) is only significant in the most challenging scenario (Rec@1) but remains stable at lower recall scores.

\begin{table}[t]
    \centering
\begin{tabular}{@{}lcccc c@{}}
\toprule
 & \multicolumn{4}{c}{COCO-All} & Zero-Shot? \\ \midrule
\textbf{Recall:} & Rec@1 & Rec@3 & Rec@5 & Rec@10 & \\ \midrule
CLIP \tiny{(ViT-B)}~\cite{radford2021learning} & .350 & \textbf{.474} & .512 & .579 & \cmark \\
\multicolumn{1}{l}{CLIP + (ours)} 
    & \textbf{.355 }
    & .468 
    & \textbf{.535 }
    & \textbf{.621 }
    & \cmark \\ 
\midrule
ALIGN \tiny{\cite{jia2021scalingvisualvisionlanguagerepresentation}} & .465 & .665 & .756 & .841 & \cmark \\ 
\multicolumn{1}{l}{ALIGN + (ours)} 
    & \textbf{.509 }
    & \textbf{.715 }
    & \textbf{.812 }
    & \textbf{.882 }
    & \cmark \\ \midrule

BEIT-3 \tiny{(COCO-Finetuned)}~\cite{wang2023image} & \textbf{.665} & \textbf{.841 }& \textbf{.891} & \textbf{.962} & \xmark \\

\multicolumn{1}{l}{ BEIT + (ours)} & .600 & .815 & .876 & .959 & \xmark \\ 

    \midrule
    \multicolumn{1}{l}{(ours, base)} & .429 & .606 & .676 & .815 & \cmark \\ 
\bottomrule
\vspace{.05em} 
\end{tabular}
    \caption{Can our method be used as a re-ranker of traditional embedding-based approaches? \textbf{We highlight those situations on which the method acts as a booster} to the original performance. We observe how the LLM common-sense knowledge helps in zero-shot scenarios. In fine-tuned embeddings (BEIT-3), the drop in performance is not very significative.}
\label{tab:coco_all_results}
\end{table}

\begin{figure}
    \centering
    \renewcommand{\arraystretch}{1.2} 
    \setlength{\tabcolsep}{5pt}       

    \resizebox{\textwidth}{!}{
    \begin{tabular}{
        >{\centering\arraybackslash}m{0.35\textwidth} |
        >{\centering\arraybackslash}m{0.2\textwidth}
        >{\centering\arraybackslash}m{0.2\textwidth}
        >{\centering\arraybackslash}m{0.2\textwidth}
    }
         Query & Rank \#1 & Rank \#2 & Rank \#3 \\
         \hline
         ``A male skateboarder that an ollie off the curb of a snowy street.'' (CLIP) &
         \includegraphics[width=\linewidth]{images/results/quali/rankings/0/000000190841.jpg} &
         \includegraphics[width=\linewidth]{images/results/quali/rankings/0/000000361506.jpg} &
         \includegraphics[width=\linewidth]{images/results/quali/rankings/0/000000171382.jpg} \\
         
         ``A male skateboarder that an ollie off the curb of a snowy street.'' (ours) & 
         \includegraphics[width=\linewidth]{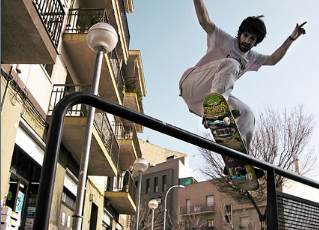} &
         \colorbox{green}{\includegraphics[width=\linewidth]{images/results/quali/rankings/0/000000253819.jpg}} &
         \includegraphics[width=\linewidth]{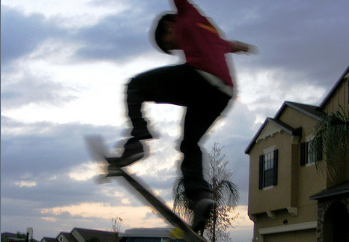} \\
                  \hline
         ``Two horses standing around in a field near a brick building'' (CLIP) &
         \includegraphics[width=\linewidth]{images/results/quali/rankings/1/000000335450.jpg} &
         \includegraphics[width=\linewidth]{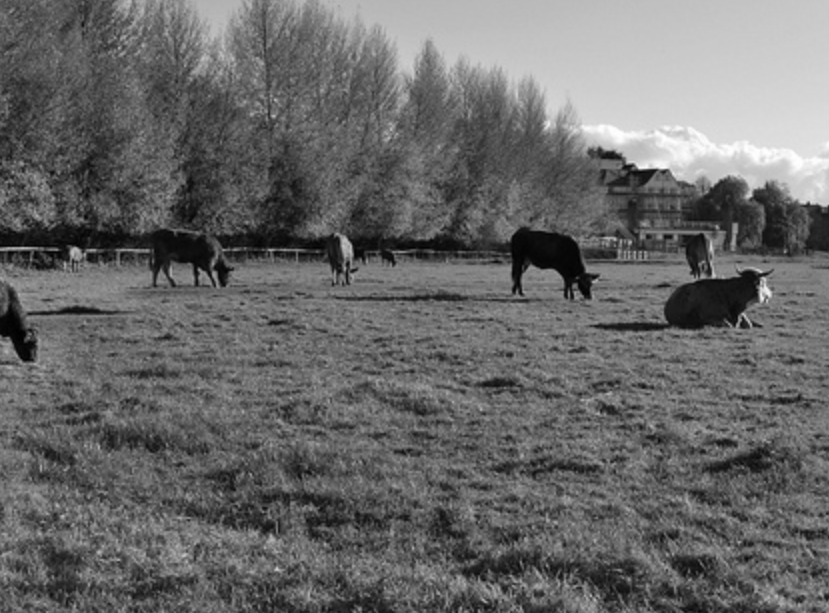} &
         \includegraphics[width=\linewidth]{images/results/quali/rankings/1/000000369081.jpg} \\
         
         ``Two horses standing around in a field near a brick building'' (ours) & 
         \colorbox{green}{\includegraphics[width=\linewidth]{images/results/quali/rankings/1/000000234807.jpg}} &
         \includegraphics[width=\linewidth]{images/results/quali/rankings/1/000000051309.jpg} &
         \includegraphics[width=\linewidth]{images/results/quali/rankings/1/000000298697.jpg} \\

                           \hline
         ``Two giraffes make contact while a third eats from a tree.'' (CLIP) &
         \includegraphics[width=\linewidth]{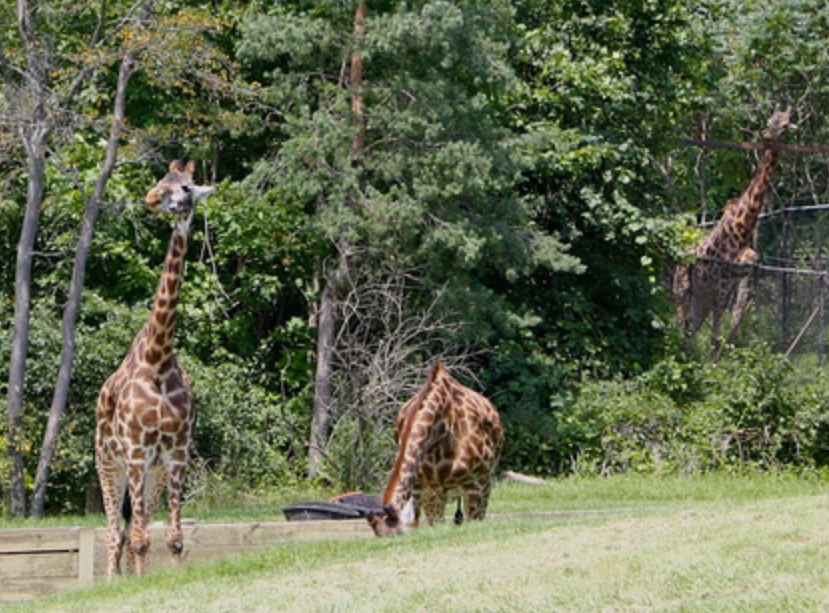} &
         \includegraphics[width=\linewidth]{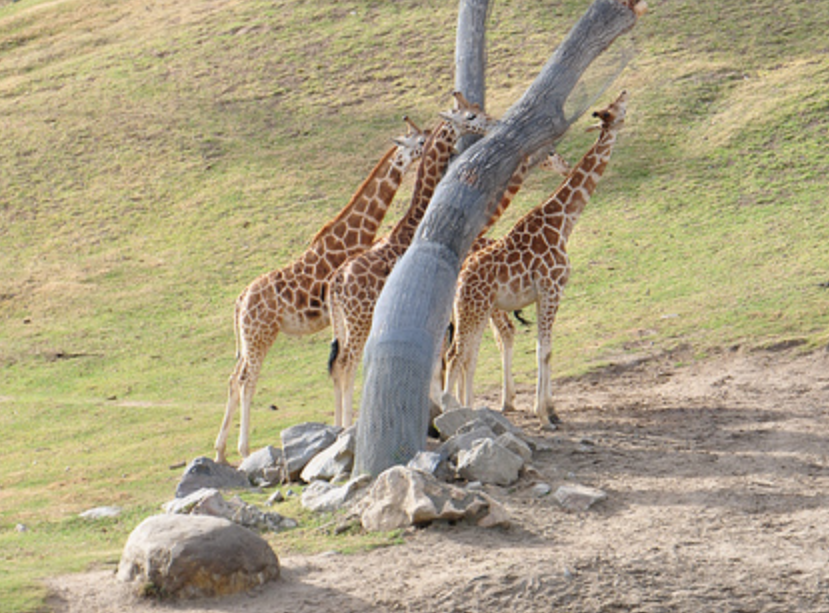} &
         \includegraphics[width=\linewidth]{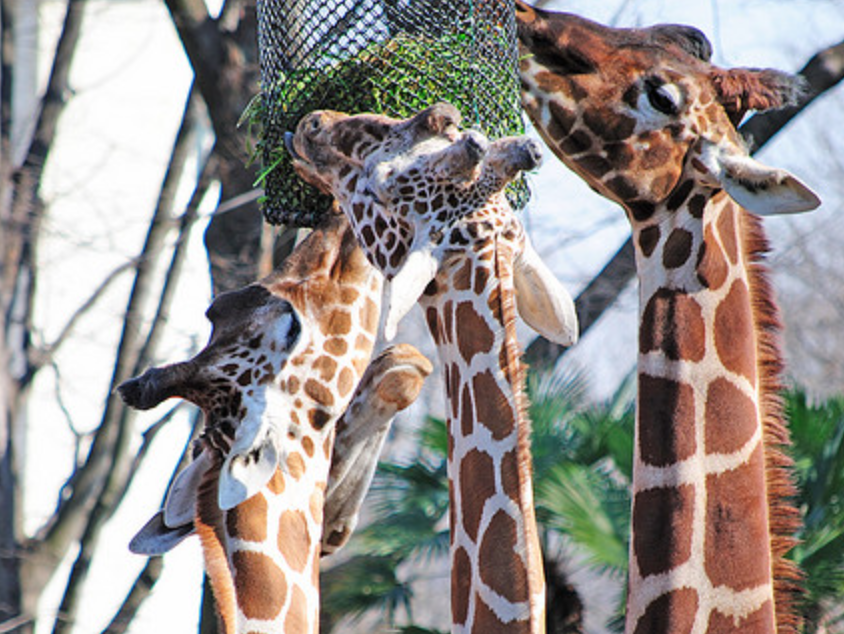} \\
         
         ``Two giraffes make contact while a third eats from a tree.'' (ours) & 
         \colorbox{green}{\includegraphics[width=\linewidth]{images/results/quali/rankings/2/000000157098.jpg}} &
         \includegraphics[width=\linewidth]{images/results/quali/rankings/2/000000165336.jpg} &
         \includegraphics[width=\linewidth]{images/results/quali/rankings/2/retall.png} \\

                           \hline
         ``A woman being photographed holding a baseball bat.'' (CLIP) &
         \colorbox{green}{\includegraphics[width=\linewidth]{images/results/quali/rankings/3/000000284743.jpg}} &
         \includegraphics[width=\linewidth]{images/results/quali/rankings/3/000000033221.jpg} &
         \includegraphics[width=\linewidth]{images/results/quali/rankings/3/000000556000.jpg} \\
         
         ``A woman being photographed holding a baseball bat.'' (ours) & 
         \colorbox{green}{\includegraphics[width=\linewidth]{images/results/quali/rankings/3/000000284743.jpg}} &
         \includegraphics[width=\linewidth]{images/results/quali/rankings/3/000000429690.jpg} &
         \includegraphics[width=\linewidth]{images/results/quali/rankings/3/000000470779.jpg} \\
         
    \end{tabular}}
    \caption{Example of retrieving information by both using CLIP and our proposed verifiable framework. Green background indicates a groundtruth image being ranked among the Top~3.}
    \label{fig:rankings}
\end{figure}

\paragraph{Qualitative Evaluation} {After qualitatively inspecting results on the COCO-Hard split, we identify three key advantages of verifiable visual search, as illustrated in \figureautorefname~\ref{fig:qualitatius}. These include the system’s ability to recognize text, perform compositional reasoning, and handle multi-faceted captions. For a more comprehensive qualitative evaluation, including failure cases and additional examples that highlight the significance of integrating contextual knowledge into visual routines.

\begin{figure}[t]
    \centering
\begin{tabular}{>{\centering\arraybackslash}m{0.48\textwidth} >{\centering\arraybackslash}m{0.48\textwidth}}
     CLIP Top-1 & Ours Top-1 \\
     \includegraphics[width=\linewidth]{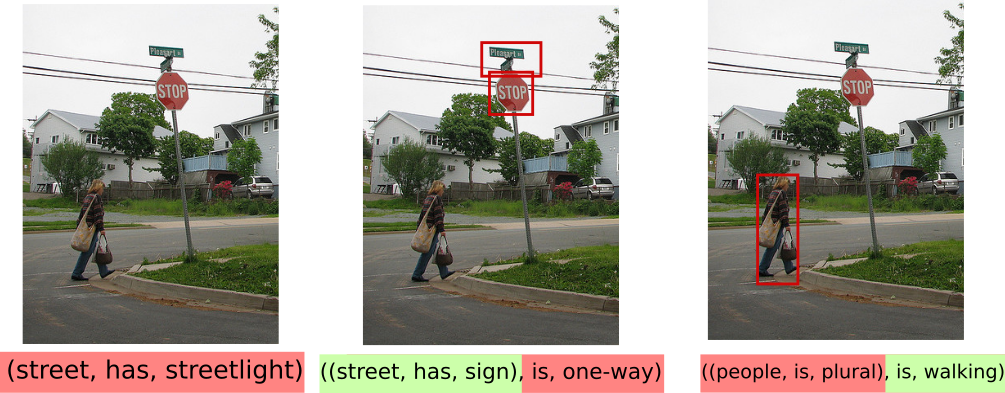} &
     \includegraphics[width=\linewidth]{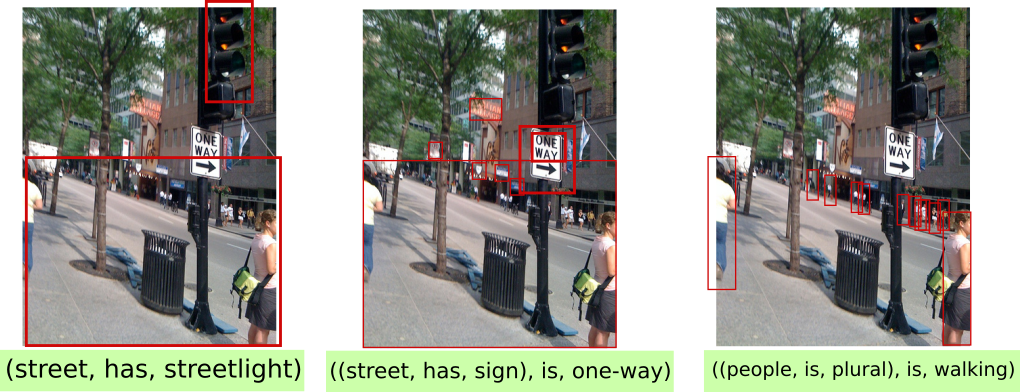} \\
     
     \includegraphics[width=\linewidth]{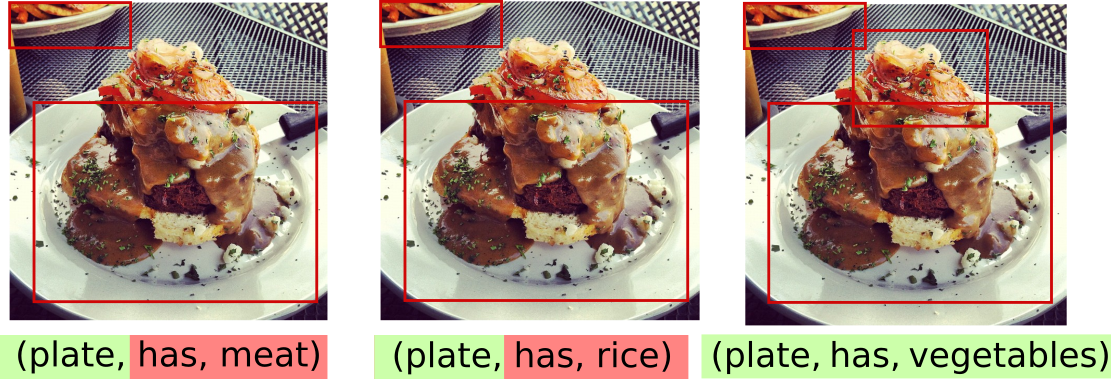} &
     \includegraphics[width=\linewidth]{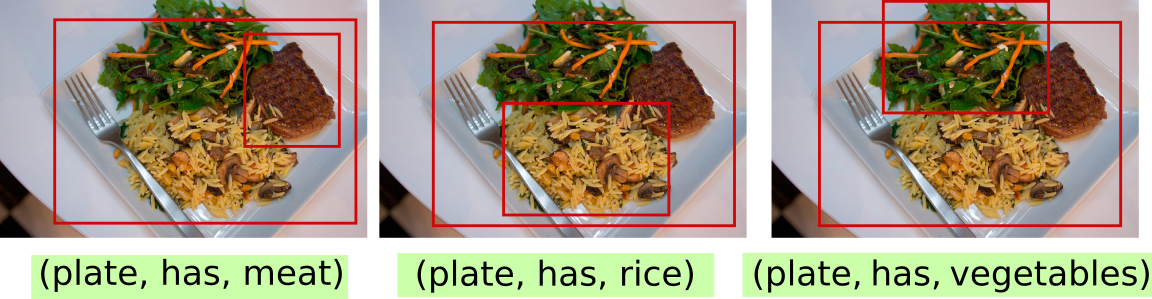} \\
     
     \includegraphics[width=\linewidth]{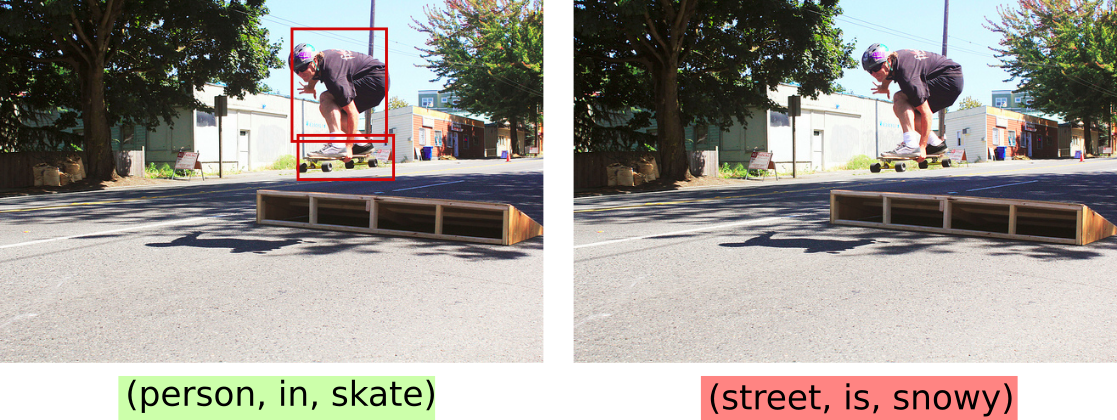} &
     \includegraphics[width=\linewidth]{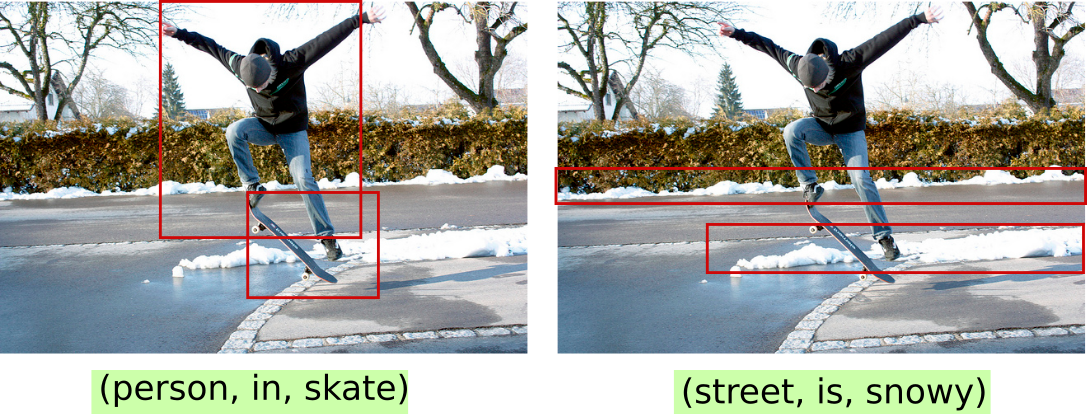} \\
\end{tabular}
\caption{Top-1 retrieved results for CLIP (left) and our method (right). The verification process in our approach helps promote the correct image by recognizing textual cues and compositional features. Notably, in cases involving multi-faceted captions (bottom row), CLIP tends to prioritize a dominant feature while overlooking secondary attributes, whereas our method demonstrates improved holistic understanding.}
\label{fig:qualitatius}
\end{figure}

First, CLIP and other embedding-based methods often struggle with the level of specificity required by text recognition. For example, while these models may retrieve images containing signs, only an explicit reading system can discern directional indicators with the necessary granularity. In the same example, our method demonstrates the ability to distinguish between ``people'' and ``person.'' While both candidate images include pedestrians, the visual routine can count them explicitly, enabling more traceable and accurate decisions.

Secondly, traditional visual search systems often struggle with complex image compositions. Such images are frequently favored because their generic visual embeddings align with a broad range of textual queries. In contrast, the compositional structure of our visual routines allows the framework to decompose an image into its fundamental components, enabling more precise matching.

Lastly, CLIP also exhibits limitations when handling multi-faceted captions: those that include two or more distinct, and potentially uncorrelated, attributes. In these cases, CLIP tends to prioritize the dominant textual feature while disregarding secondary elements. Our method, by contrast, evaluates each component of the caption independently, leading to more balanced and accurate retrieval.

Interestingly, embedding-based methods can also return reasonable results when the visual overlap aligns well with the text. However, this is not always grounded in semantic correctness. For example, in the last query shown in  as shown in \figureautorefname~\ref{fig:rankings}, our method erroneously places an image in the top-3 due to the object detector misidentifying skis as a baseball bat. Because the model interprets the woman as holding this incorrectly detected bat, the image is promoted in the ranking. A similar issue appears in CLIP’s second-ranked image, where a tennis racket may have been mistaken for a bat. However, in CLIP’s case, this remains speculative, relying on human interpretation of the model's behavior rather than any explicit rationale provided by the system. In contrast, as shown in Table~\ref{tab:res}, the errors of our visual routines are fully traceable an interpretable; with errors ranging from early stages of the pipeline (code generation) to issues regarding the code execution (under-performance of the image detection algorithm).  
}

In summary, our verification-based approach offers a competitive alternative to traditional embedding-based methods in text-to-image visual search, particularly in scenarios requiring fine-grained reasoning where compositionality, numerical reasoning, and textual understanding are necessary. Although the overall gains are moderated by noise introduced through code synthesis and visual detection APIs, our framework maintains robustness and interpretability. Qualitative analyses further confirm these strengths, demonstrating improved performance in tasks involving specific text recognition, compositional decomposition, and multi-faceted queries,underscoring the value of integrating verifiability into image retrieval systems.

\begin{table}
    \centering
    \resizebox{\textwidth}{!}{
    {\large 
    
    \begin{tabular}{
    >{\centering\arraybackslash}m{4cm}   
    >{\centering\arraybackslash}m{3.5cm} 
    >{\centering\arraybackslash}m{5cm}   
    | >{\centering\arraybackslash}m{2.5cm}
    >{\centering\arraybackslash}m{3cm}   
}
    \toprule
    Triplet ( $\varphi_i \in \varphi$ ) & Routine ( $\pi_i \in \Pi$ ) & Groundtruth Image ( $v_{gt}$ ) & $ \pi_i(v_{gt}) $ & Error Type \\
    \midrule
    $\text{bench} \xrightarrow{\text{located by}} \text{shore}$ & Code~1.5 & \includegraphics[width=.3\textwidth]{images/results/supp/example_5/000000271457.jpg} & \texttt{True} & -- \\
    $\text{woman} \xrightarrow{\text{posing}} \text{picture}$ & Code~1.6 & \includegraphics[width=.3\textwidth]{images/results/supp/example_6/000000309655.jpg} & \texttt{False} & Bad triplet \\
    $\text{lake} \xrightarrow{\text{with}} \text{two boats}$ & Code~1.7 & \includegraphics[width=.3\textwidth]{images/results/supp/example_7/000000171611.jpg} & \texttt{False} & Annotation is wrong \\
    $\text{road} \xrightarrow{\text{made of}} \text{dirt}$ & Code~1.8 & \includegraphics[width=.3\textwidth]{images/results/supp/example_8/000000047010.jpg} & \texttt{False} & Bad code execution \\
    $\text{sign} \xrightarrow{\text{reads}} \text{Norfolk}$ & Code~1.9 & \includegraphics[width=.3\textwidth]{images/results/supp/example_9/000000338718.jpg} & \texttt{True} & -- \\
    $\text{sign} \xrightarrow{\text{reads}} \text{Cambridge}$ & Code~1.10 & \includegraphics[width=.3\textwidth]{images/results/supp/example_9/000000338718.jpg} & \texttt{True} & -- \\
    $\text{snow} \xrightarrow{\text{is}} \text{under}$ &  Code~1.11 & \includegraphics[width=.3\textwidth]{images/results/supp/example_10/000000041990.jpg} & \texttt{False} & Bad triplet \\
    $\text{bathtub} \xrightarrow{\text{is}} \text{white}$ & Code~1.12 & \includegraphics[width=.3\textwidth]{images/results/supp/example_11/000000558073.jpg} & \texttt{False} & Bad code generation \\
    \bottomrule
\end{tabular}
    } 
}
    \vspace{1pt}
    \caption{Positive and negative examples and types of errors of some matches between triplets and images. We observe examples of reading, counting, and geometrical relationships.}
    \label{tab:res}
\end{table}

\section{Conclusion}
\label{sec:conc}

In this paper, we demonstrated the applicability of model checking in the context of visual search by leveraging the code generation capabilities of large language models. While occasional errors introduced during code synthesis may offset some of the advantages, the overall framework is positively validated. Our experimental setup consistently reveals comparable trends across different embedding-based baselines. Future work should focus on extending the framework to more complex and realistic scenarios, including the identification of named entities and the contextualization of their relationships within real-world events.
However, model checking offers a promising direction for interpretable and compositional visual search systems, opening opportunities for research at the intersection of vision, language, and symbolic reasoning.

\paragraph{Ethical Concerns} This method builds on the common-sense capabilities of large language models (LLMs) for constructing unbiased visual routines which are truly image-agnostic. In the context of image search involving everyday objects, no immediate ethical risks have been identified, suggesting a degree of safe and practical applicability. However, applying this framework to scenarios involving individuals, named entities, or culturally sensitive content is yet to be validated in further research. In such cases, generated visual routines may reflect the inherent biases of the underlying language model, potentially leading to incorrect or biased outputs without guarantees of fairness or generality.

\section*{Acknowledgements}
This work has been partially supported by the Spanish project PID2024-157778OB-I00, Ministerio de Ciencia e Innovación, the Departament de Cultura of the Generalitat de Catalunya, and the CERCA Program / Generalitat de Catalunya. Adrià Molina is funded with the PRE2022-101575 grant provided by MCIN / AEI / 10.13039 / 501100011033 and by the European Social Fund (FSE+). 

\small\printbibliography

@article{xu2023metaclip,
  title={Demystifying CLIP Data},
  author={Hu Xu, et.al},
   journal={arXiv preprint arXiv:2309.16671},
  year={2023},
}

@article{epstein2024can,
  title={Can biased search results change people’s opinions about anything at all? a close replication of the Search Engine Manipulation Effect (SEME)},
  author={Epstein, Robert and Li, Ji},
  journal={Plos one},
  volume={19},
  number={3},
  pages={e0300727},
  year={2024},
  publisher={Public Library of Science San Francisco, CA USA}
}

@article{ludolph2016manipulating,
  title={Manipulating Google’s knowledge graph box to counter biased information processing during an online search on vaccination: Application of a technological debiasing strategy},
  author={Ludolph, Ramona and Allam, Ahmed and Schulz, Peter J},
  journal={Journal of medical Internet research},
  volume={18},
  number={6},
  pages={e137},
  year={2016},
  publisher={JMIR Publications Toronto, Canada}
}

@article{clarke2018introduction,
  title={Introduction to model checking},
  author={Clarke, Edmund M and Henzinger, Thomas A and Veith, Helmut},
  journal={Handbook of Model Checking},
  pages={1--26},
  year={2018},
  publisher={Springer}
}

@article{giannakopoulou2018compositional,
  title={Compositional reasoning},
  author={Giannakopoulou, Dimitra and Namjoshi, Kedar S and P{\u{a}}s{\u{a}}reanu, Corina S},
  journal={Handbook of Model Checking},
  pages={345--383},
  year={2018},
  publisher={Springer}
}

@article{krishna2017visual,
  title={Visual genome: Connecting language and vision using crowdsourced dense image annotations},
  author={Krishna, Ranjay and Zhu, Yuke and Groth, Oliver and Johnson, Justin and Hata, Kenji and Kravitz, Joshua and Chen, Stephanie and Kalantidis, Yannis and Li, Li-Jia and Shamma, David A and others},
  journal={International journal of computer vision},
  volume={123},
  pages={32--73},
  year={2017},
  publisher={Springer}
}

@article{wust2024pix2code,
  title={Pix2code: Learning to compose neural visual concepts as programs},
  author={W{\"u}st, Antonia and Stammer, Wolfgang and Delfosse, Quentin and Dhami, Devendra Singh and Kersting, Kristian},
  journal={arXiv preprint arXiv:2402.08280},
  year={2024}
}

@inproceedings{ellis2021dreamcoder,
  title={Dreamcoder: Bootstrapping inductive program synthesis with wake-sleep library learning},
  author={Ellis, Kevin and Wong, Catherine and Nye, Maxwell and Sabl{\'e}-Meyer, Mathias and Morales, Lucas and Hewitt, Luke and Cary, Luc and Solar-Lezama, Armando and Tenenbaum, Joshua B},
  booktitle={Proceedings of the 42nd acm sigplan international conference on programming language design and implementation},
  pages={835--850},
  year={2021}
}

@inproceedings{suris2023vipergpt,
  title={Vipergpt: Visual inference via python execution for reasoning},
  author={Sur{\'\i}s, D{\'\i}dac and Menon, Sachit and Vondrick, Carl},
  booktitle={Proceedings of the IEEE/CVF International Conference on Computer Vision},
  pages={11888--11898},
  year={2023}
}

@inproceedings{lin2014microsoft,
  title={Microsoft coco: Common objects in context},
  author={Lin, Tsung-Yi and Maire, Michael and Belongie, Serge and Hays, James and Perona, Pietro and Ramanan, Deva and Doll{\'a}r, Piotr and Zitnick, C Lawrence},
  booktitle={Computer vision--ECCV 2014: 13th European conference, zurich, Switzerland, September 6-12, 2014, proceedings, part v 13},
  pages={740--755},
  year={2014},
  organization={Springer}
}

@article{stefanini2022show,
  title={From show to tell: A survey on deep learning-based image captioning},
  author={Stefanini, Matteo and Cornia, Marcella and Baraldi, Lorenzo and Cascianelli, Silvia and Fiameni, Giuseppe and Cucchiara, Rita},
  journal={IEEE transactions on pattern analysis and machine intelligence},
  volume={45},
  number={1},
  pages={539--559},
  year={2022},
  publisher={IEEE}
}

@article{sarto2025image,
  title={Image Captioning Evaluation in the Age of Multimodal LLMs: Challenges and Future Perspectives},
  author={Sarto, Sara and Cornia, Marcella and Cucchiara, Rita},
  journal={arXiv preprint arXiv:2503.14604},
  year={2025}
}

@inproceedings{radford2021learning,
  title={Learning transferable visual models from natural language supervision},
  author={Radford, Alec and Kim, Jong Wook and Hallacy, Chris and Ramesh, Aditya and Goh, Gabriel and Agarwal, Sandhini and Sastry, Girish and Askell, Amanda and Mishkin, Pamela and Clark, Jack and others},
  booktitle={International conference on machine learning},
  pages={8748--8763},
  year={2021},
  organization={PmLR}
}

@article{minderer2023scaling,
  title={Scaling open-vocabulary object detection},
  author={Minderer, Matthias and Gritsenko, Alexey and Houlsby, Neil},
  journal={Advances in Neural Information Processing Systems},
  volume={36},
  pages={72983--73007},
  year={2023}
}

@article{abdin2024phi,
  title={Phi-4 technical report},
  author={Abdin, Marah and Aneja, Jyoti and Behl, Harkirat and Bubeck, S{\'e}bastien and Eldan, Ronen and Gunasekar, Suriya and Harrison, Michael and Hewett, Russell J and Javaheripi, Mojan and Kauffmann, Piero and others},
  journal={arXiv preprint arXiv:2412.08905},
  year={2024}
}

@inproceedings{gomez2017self,
  title={Self-supervised learning of visual features through embedding images into text topic spaces},
  author={Gomez, Lluis and Patel, Yash and Rusinol, Mar{\c{c}}al and Karatzas, Dimosthenis and Jawahar, CV},
  booktitle={Proceedings of the ieee conference on computer vision and pattern recognition},
  pages={4230--4239},
  year={2017}
}

@misc{zhai2023sigmoidlosslanguageimage,
      title={Sigmoid Loss for Language Image Pre-Training}, 
      author={Xiaohua Zhai and Basil Mustafa and Alexander Kolesnikov and Lucas Beyer},
      year={2023},
      eprint={2303.15343},
      archivePrefix={arXiv},
      primaryClass={cs.CV},
      url={https://arxiv.org/abs/2303.15343}, 
}

@inproceedings{wang2023image,
  title={Image as a foreign language: Beit pretraining for vision and vision-language tasks},
  author={Wang, Wenhui and Bao, Hangbo and Dong, Li and Bjorck, Johan and Peng, Zhiliang and Liu, Qiang and Aggarwal, Kriti and Mohammed, Owais Khan and Singhal, Saksham and Som, Subhojit and others},
  booktitle={Proceedings of the IEEE/CVF Conference on Computer Vision and Pattern Recognition},
  pages={19175--19186},
  year={2023}
}

@inproceedings{jia2021scaling,
  title={Scaling up visual and vision-language representation learning with noisy text supervision},
  author={Jia, Chao and Yang, Yinfei and Xia, Ye and Chen, Yi-Ting and Parekh, Zarana and Pham, Hieu and Le, Quoc and Sung, Yun-Hsuan and Li, Zhen and Duerig, Tom},
  booktitle={International conference on machine learning},
  pages={4904--4916},
  year={2021},
  organization={PMLR}
}

@inproceedings{chun2021probabilistic,
  title={Probabilistic embeddings for cross-modal retrieval},
  author={Chun, Sanghyuk and Oh, Seong Joon and De Rezende, Rafael Sampaio and Kalantidis, Yannis and Larlus, Diane},
  booktitle={Proceedings of the IEEE/CVF conference on computer vision and pattern recognition},
  pages={8415--8424},
  year={2021}
}

@article{chun2023improved,
  title={Improved probabilistic image-text representations},
  author={Chun, Sanghyuk},
  journal={arXiv preprint arXiv:2305.18171},
  year={2023}
}

@article{wang2024balanced,
  title={Balanced similarity with auxiliary prompts: towards alleviating text-to-image retrieval bias for CLIP in zero-shot learning},
  author={Wang, Hanyao and Zhan, Yibing and Liu, Liu and Ding, Liang and Yu, Jun},
  journal={arXiv e-prints},
  pages={arXiv--2402},
  year={2024}
}

@misc{jia2021scalingvisualvisionlanguagerepresentation,
      title={Scaling Up Visual and Vision-Language Representation Learning With Noisy Text Supervision}, 
      author={Chao Jia and Yinfei Yang and Ye Xia and Yi-Ting Chen and Zarana Parekh and Hieu Pham and Quoc V. Le and Yunhsuan Sung and Zhen Li and Tom Duerig},
      year={2021},
      eprint={2102.05918},
      archivePrefix={arXiv},
      primaryClass={cs.CV},
      url={https://arxiv.org/abs/2102.05918}, 
}

@inproceedings{gupta2023visual,
  title={Visual programming: Compositional visual reasoning without training},
  author={Gupta, Tanmay and Kembhavi, Aniruddha},
  booktitle={Proceedings of the IEEE/CVF conference on computer vision and pattern recognition},
  pages={14953--14962},
  year={2023}
}

\clearpage
\end{document}